\documentclass[11pt]{article}
\usepackage[utf8]{inputenc}
\DeclareUnicodeCharacter{03B1}{\ensuremath{\alpha}}
\usepackage{times}
\usepackage{graphicx}
\usepackage{amsmath, amssymb}
\usepackage{hyperref}
\usepackage{natbib}
\usepackage{listings}
\usepackage{booktabs}
\usepackage{multirow}
\usepackage{graphicx} 
\usepackage{float} 
\usepackage{subcaption}
\usepackage{placeins} 
\usepackage[skip=2pt]{caption}
\usepackage[final]{neurips_mlsb_2025}
\usepackage[table]{xcolor}

\title{TangledFeatures: Robust Feature Selection in Highly Correlated Spaces}

\author{Allen Daniel Sunny \\
University of Maryland, College Park \\
\texttt{allens@umd.edu}}
\date{}

\begin{document}

\maketitle

\begin{abstract}

Feature selection is a fundamental step in model development, shaping both predictive performance and interpretability. Yet, most widely used methods focus on predictive accuracy, and their performance degrades in the presence of correlated predictors. To address this gap, we introduce TangledFeatures, a framework for feature selection in correlated feature spaces. It identifies representative features from groups of entangled predictors, reducing redundancy while retaining explanatory power. The resulting feature subset can be directly applied in downstream models, offering a more interpretable and stable basis for analysis compared to traditional selection techniques. We demonstrate the effectiveness of TangledFeatures on Alanine Dipeptide, applying it to the prediction of backbone torsional angles $\phi$ and $\psi$, and show that the selected features correspond to structurally meaningful intra-atomic distances that explain variation in these angles.
\end{abstract}

\section{Introduction}

Predictive modeling in structural biology has advanced rapidly, moving from early statistical methods to deep learning systems such as AlphaFold \cite{jumper_highly_2021}, which achieves remarkable accuracy in protein structure prediction. Yet, there is a gap in understanding the drivers behind these predictions: specific residues, motifs, or the correct structural interactions that determine outcomes \cite{medina-ortiz_interpretable_2025}.

For features identified as drivers to be meaningful, they must meet two goals. First, they should be biologically interpretable, mapping back to known structural or functional elements \cite{ahlquist_enabling_2023}. Second, they should be reproducible, consistently emerging across analyses rather than reflecting noise or model instability \cite{braghetto_interpretable_2023}. Meeting these goals is essential not only for scientific trust, but also for actionable insights, enabling mutational studies, protein design, and functional discovery \cite{chen_applying_2024}.

Approaches to interpretability in structural biology fall under two categories: post-hoc and pre-hoc. Post-hoc methods, such as SHAP \cite{lundberg_unified_2017}, Integrated Gradients \cite{sundararajan_axiomatic_2017}, and Global Importance Analysis \cite{koo_global_nodate} apply explanatory frameworks after training. However, explanations often unstable in the presence of highly correlated structural features and may fail to yield biologically meaningful drivers \cite{molnar_general_2021}. Pre-hoc approaches have attempted to mitigate this by designing interpretable feature spaces before modeling. For example, clustering residue–residue contacts from molecular dynamics prior to Random Forest classification \cite{callahan_interpretable_nodate}, or using sparse autoencoders to extract latent motifs from protein language models \cite{lyons_predicting_2014}. While these strategies reduce redundancy, they still exhibit instability. For example, different runs frequently produce divergent feature sets, leaving reproducibility an open challenge. Unlike earlier clustering-based pipelines, our method explicitly evaluates feature stability across runs, ensuring that identified drivers are not only interpretable but also reproducible.

To address the instability of existing interpretability methods, we place feature stability at the center of our approach. Our objectives are: (i) to evaluate the reproducibility of structural drivers across multiple analyses; (ii) to distinguish robust biological determinants from spurious correlations; and (iii) to provide a systematic framework for actionable biochemical insight.

\begin{figure}[ht]
    \centering
    \includegraphics[width=0.6\linewidth]{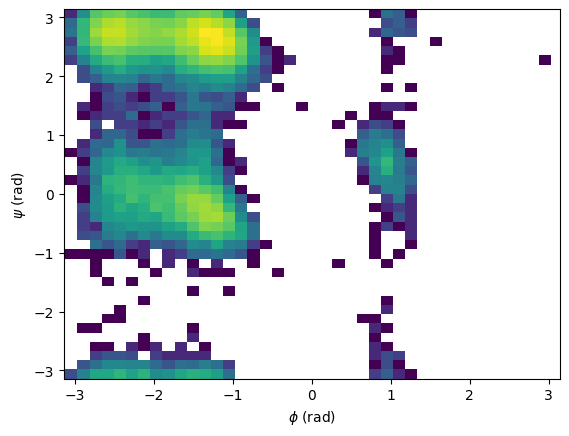}
    \caption{Illustration of the Ramachandran plot, showing allowed conformational regions for the backbone torsional angles $\phi$ and $\psi$. 
    In this work, $\phi$ (C–N–C$_\alpha$–C) and $\psi$ (N–C$_\alpha$–C–N) serve as prediction targets, while intra-atomic distances are used as input features.}
    \label{fig:ramachandran}
\end{figure}

To ground our study, we focus on the canonical case of \textit{Alanine dipeptide}, a widely used benchmark in structural biology \cite{feig_is_2008}. Our prediction targets are the backbone torsional angles $\phi$ (C–N–C$_\alpha$–C) and $\psi$ (N–C$_\alpha$–C–N), with intra-atomic distances serving as the input features. This setup allows us to rigorously test whether our framework can identify stable and biologically meaningful drivers of conformational variation \cite{kikutsuji_explaining_2022} (see Fig. ~\ref{fig:ramachandran}). 


Our contributions are threefold. First, we introduce TangledFeatures, a stability-based pipeline for feature selection in correlated spaces. Second, we validate it quantitatively through predictive accuracy and stability analyses. Third, we demonstrate its interpretability by showing that selected features correspond to meaningful intra-atomic distances in Alanine Dipeptide.
Our results reveal TangledFeaures successfully maps cause and effect in complex correlated systems, presenting a novel and principled approach to feature selection, supporting reproducible research in structural biology.

\section{Methods}

\subsection{Overview}

Our objective is to develop a generalized feature selection algorithm that balances two central goals: preserving predictive accuracy while maintaining feature interpretability. As a proof-of-concept, we apply the algorithm to Alanine Dipeptide (Ace–Ala–Nme), a minimal peptide with a central alanine residue flanked by an acetyl group and an N-methylamide group. Its backbone conformation is defined by two torsional angles, $\phi$ (C–N–C$_\alpha$–C) and $\psi$ (N–C$_\alpha$–C–N), which serve as the prediction targets. Each conformation is represented by a vector of intra-atomic distances among heavy atoms, yielding a feature space $d \in \mathbb{R}^m$, where $m$ the number of intra-atomic distances. The task is to map $d$ to torsional angles $(\phi, \psi)$ and to evaluate whether the algorithm can isolate a compact subset of distances that captures torsional variability while preserving predictive performance.

In this section, we focus on three components of the TangledFeatures pipeline. The correlation module $c_\alpha$ computes pairwise correlations and constructs a graph $G$, where nodes represent distances and edges denote correlations above a threshold. The selection module $s_\beta$ operates on each connected component of $G$ to extract a representative feature, thereby eliminating redundancy. The refinement module $r_\gamma$ applies random forest feature selection to the set of representatives, filtering them further by predictive importance to produce the final subset of variables.

The overall process can be summarized as:
\[
d \xrightarrow{c_\alpha} d_c \xrightarrow{s_\beta} d_s \xrightarrow{r_\gamma} d' \xrightarrow{f} (\phi, \psi).
\]

\subsection{Clustering}
The first stage of TangledFeatures is to identify groups of highly correlated features while ensuring the prediction targets are treated consistently. Given the feature matrix $D \in \mathbb{R}^{n \times m}$, where $n$ is the number of conformations and $m$ the number of intra-atomic distances, we compute the pairwise Pearson correlation matrix $\Sigma \in \mathbb{R}^{m \times m}$. An undirected graph $G = (V,E)$ is then constructed with vertices $V$ corresponding to features and edges 
\[
E = \{(i,j) : |\Sigma_{ij}| \geq \tau\},
\] 
where $\tau$ is a user-specified threshold. Connected components of $G$ define clusters of features that convey largely redundant information. To account for the periodicity of torsional angles, the prediction targets $(\phi, \psi)$ are represented in cosine–sine form $(\cos \phi, \sin \phi, \cos \psi, \sin \psi)$, which removes discontinuities at angular boundaries and ensures stable downstream analysis.

\subsection{Selection Module}
For each correlated cluster $C_k$, the selection module $s_\beta$ identifies a single representative distance using an ensemble-based stability selection procedure. Following the principle of stability selection \citep{kalousis_stability_2007}, and ensemble feature selection methods such as Boruta \cite{kursa_boruta_2010}, and Random Forest variable importance measures \cite{louppe_understanding_nodate}, we train multiple random forests where, in each run, one candidate distance is sampled from each cluster and evaluated alongside all uncorrelated variables when predicting the torsional targets $(\cos \phi, \sin \phi, \cos \psi, \sin \psi)$. We average feature importance scores across runs,
\[
\hat{I}(d_{ij}) = \frac{1}{R}\sum_{r=1}^R I_r(d_{ij}),
\]
and the distance with the highest $\hat{I}$ within each cluster is retained as the representative. This ensemble approach reduces variance and prevents arbitrary choices among correlated distances, ensuring that the final representatives consistently capture the structural drivers of torsional variability.

\subsection{Refinement Module}
After cluster representatives are identified, the refinement module $r_\gamma$ applies a Random Forest feature selection step based on cumulative importance coverage. The representatives are ranked by their Random Forest importance scores, and features are retained in descending order until the cumulative importance reaches 0.99 \cite{louppe_understanding_nodate}. This threshold ensures that the final subset $d'$ captures at least 99\% of the predictive signal while discarding variables that contribute minimally. By combining local stability within clusters and global coverage across the model, the refinement module yields a compact set of distances that preserves predictive accuracy without sacrificing interpretability.

\subsection{Objective} \label{subsec:objective}

\textbf{Feature Stability:} To assess the robustness of selected variables, we repeated the entire pipeline across multiple resampled training sets. We quantified feature importance using SHAP values for each run \cite{jethani_fastshap_2022}. We then evaluated stability in two complementary ways. First, we computed the Spearman rank correlation \cite{de_winter_comparing_2016} between runs to capture the consistency of feature importance rankings across resamples. Second, we applied the Kuncheva index \cite{kuncheva_stability_nodate}, which compares the overlap of the top-k most important features across runs, adjusted for chance. Together, these measures assess whether both the relative ordering of features and the membership of the most important subsets remain stable under data perturbations \cite{haury_influence_2011}. A high stability score indicates that the algorithm consistently highlights the same core drivers of torsional variability \cite{kalousis_stability_2007}.

\textbf{Predictive Accuracy:} To evaluate whether interpretability is preserved without sacrificing performance, we trained predictive models (OLS, Random Forest \cite{wright_ranger_2017}, XGBoost \cite{chen_xgboost_2016}, and SVM \cite{708428}) on the selected subsets to predict torsional targets $(\cos \phi, \sin \phi, \cos \psi, \sin \psi)$. We measured accuracy on held-out conformations using root mean squared error (RMSE) and the coefficient of determination ($R^2$). The refinement module ensures that retained features capture at least 99\% of the cumulative Random Forest importance, promoting compactness while maintaining predictive fidelity.

\section{Experiments}

We evaluated TangledFeatures against popular feature selection methods on the Alanine Dipeptide system: LASSO, Elastic Net Regularization (ENR), Random Forest Recursive Feature Evaluation (RFE), Boruta and a no-selection baseline. Each feature selection method was paired with four predictive algorithms: Ordinary Least Squares (OLS), Random Forest Regression, XGBoost, and Support Vector Machines (SVM) — yielding 24 pipelines in total (see Table. \ref{tab:accuracy_results}). For each pipeline, models were trained on 80\% of the data and evaluated on the remaining 20\%. To assess robustness, we repeat the entire training and evaluation process $10$ times with independent bootstrap resamples of the training data. We evaluated along three axes: predictive accuracy, stability and interpretability.

\subsection{Predictive Accuracy}

\noindent \textbf{Metrics:} We evaluated predictive performance by training models to map intra-atomic distances to torsional targets in cosine–sine form $(\cos \phi, \sin \phi, \cos \psi, \sin \psi)$. For each pipeline, we measured accuracy on a held-out test set using root mean squared error (RMSE) and the coefficient of determination ($R^2$). We computed metrics separately for $\phi$ and $\psi$.

\textbf{Results:} Results are summarized in Table \ref{tab:accuracy_results}. All models achieve high predictive accuracy ($R^2 > 0.9$ for most pipelines). The best performance for $\phi$ is achieved by SVM with Elastic Net (RMSE = 0.05, $R^2$ = 0.99), while the best performance for $\psi$ is also obtained with SVM + Elastic Net (RMSE = 0.84, $R^2$ = 0.89). Higher accuracy for SVM arises from leveraging redundant features, whereas TangledFeatures trades a slight loss in accuracy for a more stable, non-redundant feature set.

\begin{table}[H]
\centering
\resizebox{\textwidth}{!}{
\begin{tabular}{lcccccccc}
\toprule
 & \multicolumn{4}{c}{$\phi$} & \multicolumn{4}{c}{$\psi$} \\
\cmidrule(lr){2-5} \cmidrule(lr){6-9}
 & OLS & RF & XGBoost & SVM & OLS & RF & XGBoost & SVM \\
\midrule
No Feature (baseline)   & 0.19 / 0.93 & 0.29 / 0.83 & 0.08 / 0.98 & 0.06 / 0.99 & 0.65 / 0.71 & 0.69 / 0.80 & 0.67 / 0.80 & 0.83 / 0.69 \\
LASSO                   & 0.21 / 0.92 & 0.09 / 0.98 & 0.09 / 0.98 & 0.05 / 0.99 & 0.89 / 0.64 & 0.69 / 0.79 & 0.63 / 0.82 & 0.84 / 0.70 \\
Elastic Net (ENR)       & 0.20 / 0.92 & 0.07 / 0.98 & 0.07 / 0.99 & \textbf{0.05 / 0.99} & 0.90 / 0.64 & 0.66 / 0.81 & 0.67 / 0.81 & \textbf{0.87 / 0.89} \\
RF Recursive Eval (RFE) & 0.30 / 0.83 & 0.10 / 0.97 & 0.10 / 0.97 & 0.05 / 0.99 & 0.65 / 0.81 & 0.67 / 0.81 & 0.65 / 0.82 & 0.89 / 0.67 \\
Boruta                  & 0.22 / 0.91 & 0.07 / 0.98 & 0.07 / 0.98 & 0.05 / 0.99 & 0.91 / 0.64 & 0.65 / 0.82 & 0.91 / 0.64 & 0.84 / 0.70 \\
\rowcolor{gray!10} TangledFeatures         & 0.26 / 0.87 & 0.09 / 0.98 & 0.09 / 0.97 & 0.09 / 0.98 & 0.97 / 0.61 & 0.67 / 0.81 & 0.97 / 0.60 & 0.86 / 0.75 \\
\bottomrule
\end{tabular}
} 
\vspace{0.25em} 

\caption{Predictive accuracy for torsional angle prediction. Rows list feature selection methods, while columns show predictive algorithms (OLS, RF, XGBoost, SVM) for each torsional angle $\phi$ and $\psi$. We report results as RMSE / $R^2$. The best overall results for $\phi$ and $\psi$ are shown in bold. We highlight the TangledFeatures row for comparison. TangledFeatures remains competitive with other selection methods while pruning redundancy, demonstrating that compact, interpretable subsets can be obtained without substantial loss of predictive fidelity.}
\label{tab:accuracy_results}
\end{table}
\FloatBarrier

\vspace{-1em} 
\subsection{Stability}
\textbf{Metrics:} We quantified stability using the Kuncheva index and Spearman rank correlation, as described in Section \ref{subsec:objective}. We computed metrics separately for torsional angles $\phi$ and $\psi$.

\begin{figure}[H]
    \centering
    
    \begin{subfigure}{0.45\linewidth}
        \centering
        \includegraphics[width=\linewidth]{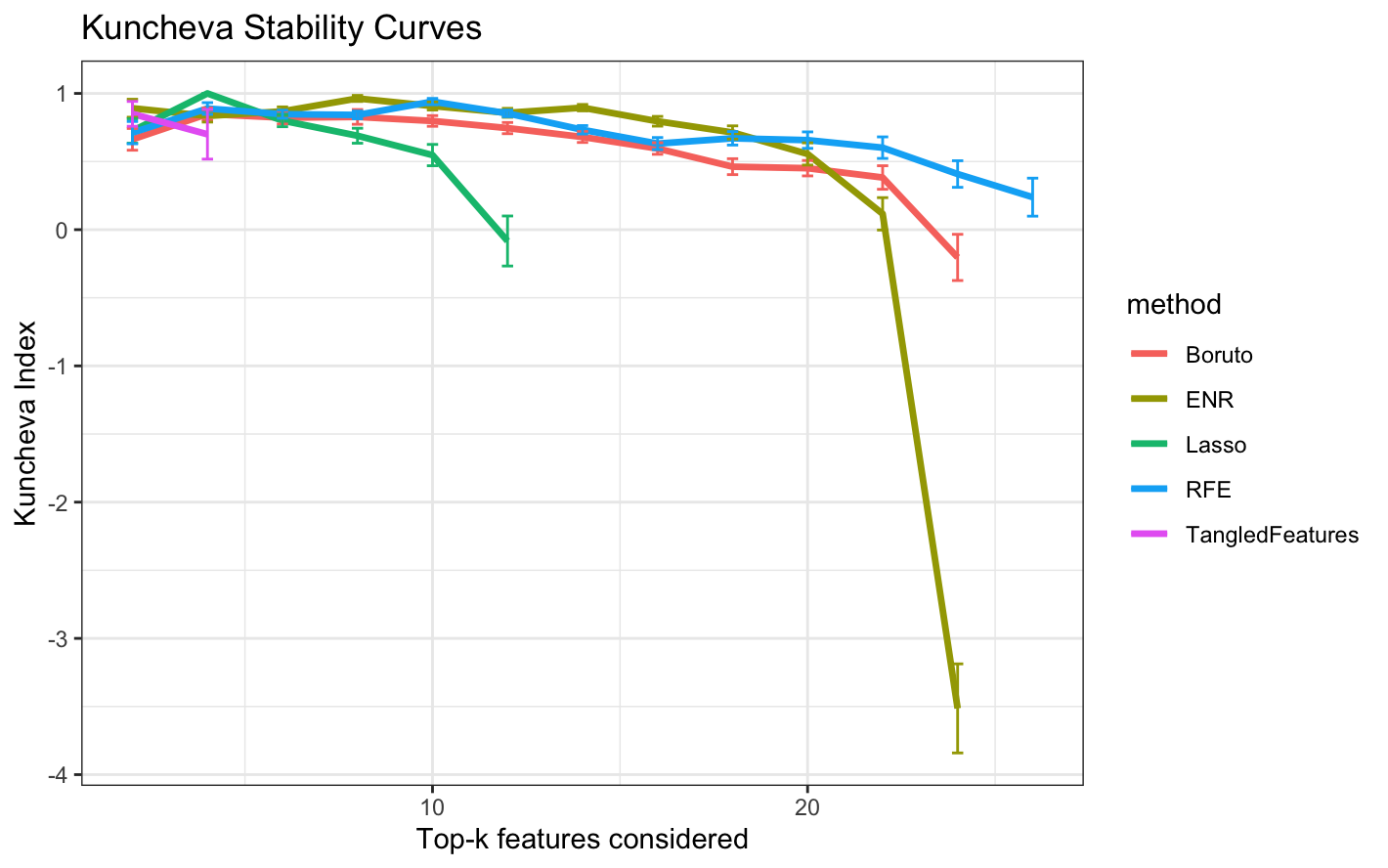}
        \caption{$\phi$ Kuncheva}
    \end{subfigure}
    \begin{subfigure}{0.45\linewidth}
        \centering
        \includegraphics[width=\linewidth]{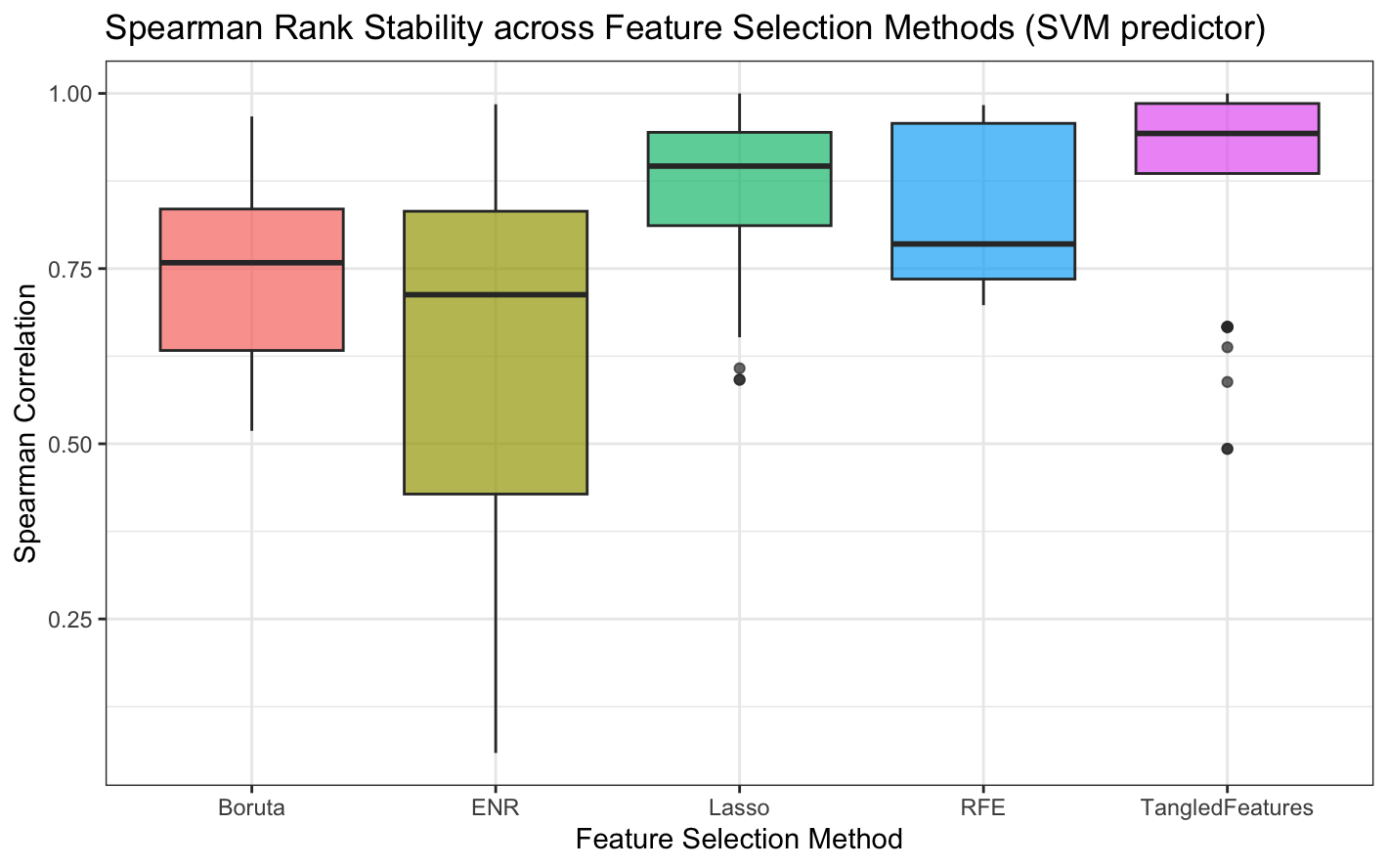}
        \caption{$\phi$ Spearman}
    \end{subfigure}

\caption{Stability for the $\phi$ torsional angle. (a) Kuncheva index, (b) Spearman rank correlation. }
\label{fig:2a}

\end{figure}

\textbf{Results:} Figures~\ref{fig:2a}a–b report stability for the $\phi$ torsional angle, based on SHAP-derived feature importance. Under the Kuncheva index (panel a), TangledFeatures achieves consistently higher overlap of the top-$k$ SHAP features across bootstrap runs than ENR or RFE. ENR shows sharp drops in overlap 
when correlated predictors are present, while RFE often retains different redundant distances, leading to low reproducibility. In contrast, TangledFeatures curves remain flat and close to the maximum, indicating nearly identical feature subsets across resamples. The Spearman correlation results (panel b) reveal the same trend at the level of feature rankings. ENR and RFE correlations fluctuate strongly across runs, reflecting instability in the relative importance assigned to correlated predictors. TangledFeatures achieves near-perfect rank correlation across resamples, showing that both the selected subset and the relative ordering of features are 
highly reproducible. For a more detailed analysis, including results for $\psi$, we refer the reader to the Appendix \ref{fig:phi_stability_grid}, \ref{fig:psi_stability_grid}.

\subsection{Interpretability}

\begin{figure}[ht]
    \centering
    \begin{subfigure}[t]{0.45\linewidth}
        \centering
        \includegraphics[width=\linewidth]{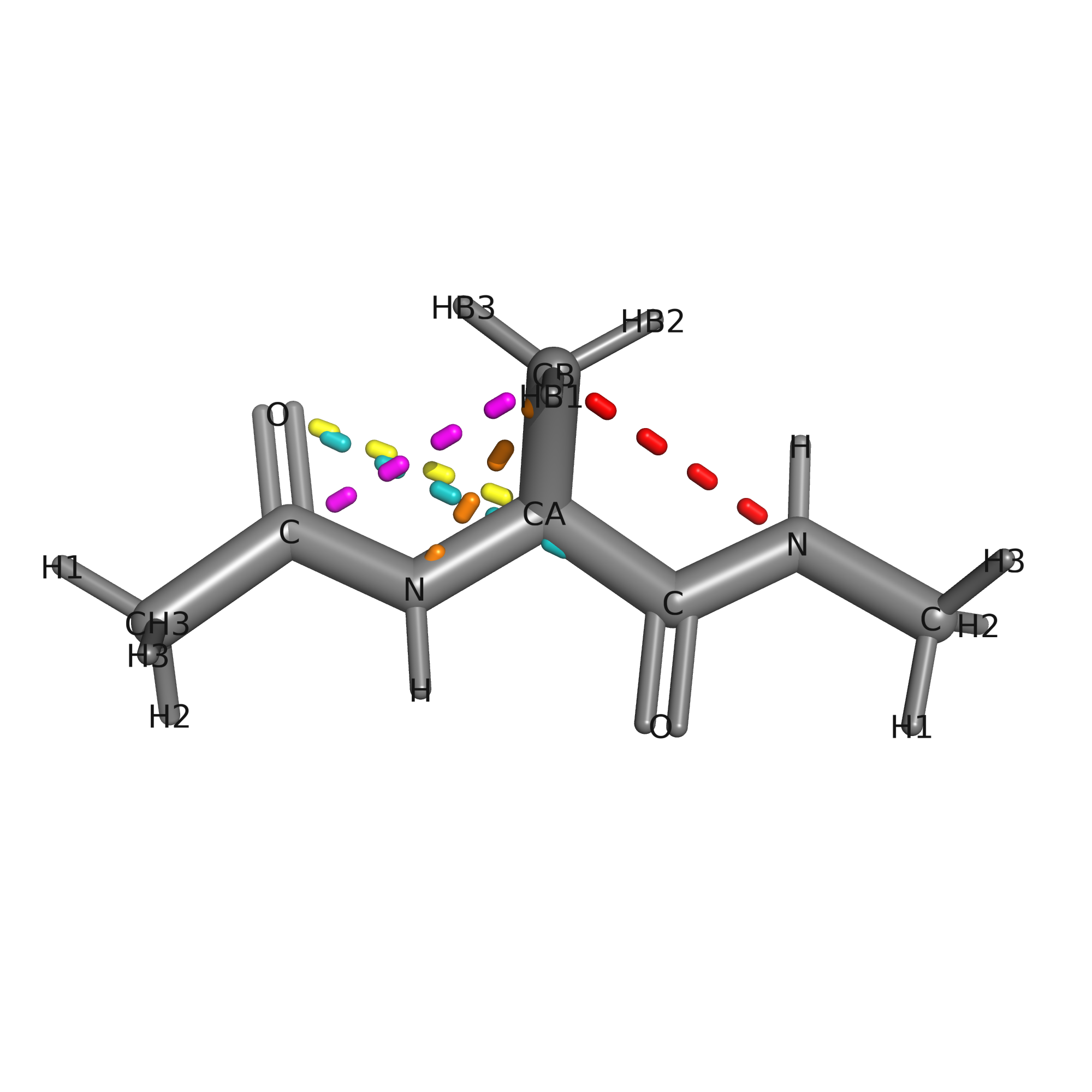}
        \caption{TangledFeatures: $\phi$ drivers}
        \label{fig:phi_tangled}
    \end{subfigure}
    \hfill
    \begin{subfigure}[t]{0.45\linewidth}
        \centering
        \includegraphics[width=\linewidth]{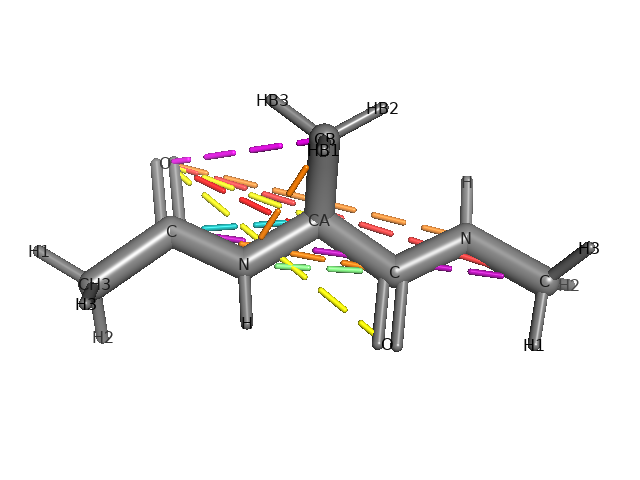}
        \caption{LASSO: $\phi$ drivers}
        \label{fig:phi_lasso}
    \end{subfigure}

    \caption{Comparison of selected feature subsets for the $\phi$ torsional angle.}
    \label{fig:interpretability}
\end{figure}

To assess interpretability, we projected the features selected by TangledFeatures onto the Alanine Dipeptide structure. The retained distances consistently aligned with backbone and near-backbone interactions—such as N–C, C–C, and C–N that are well-established determinants of $\phi$ and $\psi$ torsional variability \cite{beauchamp_are_2012} \cite{piana_assessing_2014} \cite{ksenofontova_conformational_2013}. Figure~\ref{fig:interpretability} highlights these distances on the atomic diagram of Alanine Dipeptide. As an additional comparison, we also visualize the subset obtained from LASSO, the feature selection method that retained the second-fewest features among baselines. Unlike TangledFeatures, LASSO often selects redundant or chemically less meaningful distances, limiting interpretability despite its compactness. Taken together, these results demonstrate that TangledFeatures produces subsets that are not only stable and predictive, but also consistent with structural biology knowledge of peptide backbone flexibility. A full list of selected distances can be found in the Appendix \ref{appendix:selected-distances}.

\section{Conclusion}
These experiments demonstrate that TangledFeatures excels when datasets contain structured correlations among predictors, a setting where classical methods may either oversimplify (e.g., PCA \cite{gewers_principal_2022}) or retain redundant features (e.g., LASSO, Boruta). Nevertheless, in cases where predictive accuracy is the sole objective and interpretability is less critical, PCA or LASSO may remain competitive alternatives. For the community at large, TangledFeatures provides a principled way to reconcile interpretability with robustness, ensuring that feature selection reflects the true informational diversity of the data.  

\section*{Acknowledgments}
The author gratefully thanks Xinyu Gu and Dr.~Pratyush Tiwary (University of Maryland, College Park) for providing early access to the dataset and an accompanying image used in this work, as well as for helpful preliminary feedback during the development of \textit{TangledFeatures}.

\bibliography{references}

\begin{thebibliography}{10}

\bibitem{jumper_highly_2021}
John Jumper, Richard Evans, Alexander Pritzel, Tim Green, Michael Figurnov, Olaf Ronneberger, Kathryn Tunyasuvunakool, Russ Bates, Augustin Žídek, Anna Potapenko, Alex Bridgland, Clemens Meyer, Simon A.~A. Kohl, Andrew~J. Ballard, Andrew Cowie, Bernardino Romera-Paredes, Stanislav Nikolov, Rishub Jain, Jonas Adler, Trevor Back, Stig Petersen, David Reiman, Ellen Clancy, Michal Zielinski, Martin Steinegger, Michalina Pacholska, Tamas Berghammer, Sebastian Bodenstein, David Silver, Oriol Vinyals, Andrew~W. Senior, Koray Kavukcuoglu, Pushmeet Kohli, and Demis Hassabis.
\newblock Highly accurate protein structure prediction with {AlphaFold}.
\newblock {\em Nature}, 596(7873):583--589, August 2021.

\bibitem{medina-ortiz_interpretable_2025}
David Medina-Ortiz, Ashkan Khalifeh, Hoda Anvari-Kazemabad, and Mehdi~D. Davari.
\newblock Interpretable and explainable predictive machine learning models for data-driven protein engineering.
\newblock {\em Biotechnology Advances}, 79:108495, March 2025.

\bibitem{ahlquist_enabling_2023}
K.~D. Ahlquist, Lauren~A. Sugden, and Sohini Ramachandran.
\newblock Enabling interpretable machine learning for biological data with reliability scores.
\newblock {\em PLOS Computational Biology}, 19(5):e1011175, May 2023.

\bibitem{braghetto_interpretable_2023}
Anna Braghetto, Enzo Orlandini, and Marco Baiesi.
\newblock Interpretable {Machine} {Learning} of {Amino} {Acid} {Patterns} in {Proteins}: {A} {Statistical} {Ensemble} {Approach}.
\newblock {\em Journal of Chemical Theory and Computation}, 19(17):6011--6022, September 2023.

\bibitem{chen_applying_2024}
Valerie Chen, Muyu Yang, Wenbo Cui, Joon~Sik Kim, Ameet Talwalkar, and Jian Ma.
\newblock Applying interpretable machine learning in computational biology — pitfalls, recommendations and opportunities for new developments.
\newblock {\em Nature methods}, 21(8):1454--1461, August 2024.

\bibitem{lundberg_unified_2017}
Scott Lundberg and Su-In Lee.
\newblock A {Unified} {Approach} to {Interpreting} {Model} {Predictions}, November 2017.
\newblock arXiv:1705.07874 [cs].

\bibitem{sundararajan_axiomatic_2017}
Mukund Sundararajan, Ankur Taly, and Qiqi Yan.
\newblock Axiomatic {Attribution} for {Deep} {Networks}, June 2017.
\newblock arXiv:1703.01365 [cs].

\bibitem{koo_global_nodate}
Peter~K Koo, Matthew Ploenzke, Praveen Anand, and Steffan~B Paul.
\newblock Global {Importance} {Analysis}: {A} {Method} to {Quantify} {Importance} of {Genomic} {Features} in {Deep} {Neural} {Networks}.

\bibitem{molnar_general_2021}
Christoph Molnar, Gunnar König, Julia Herbinger, Timo Freiesleben, Susanne Dandl, Christian~A. Scholbeck, Giuseppe Casalicchio, Moritz Grosse-Wentrup, and Bernd Bischl.
\newblock General {Pitfalls} of {Model}-{Agnostic} {Interpretation} {Methods} for {Machine} {Learning} {Models}, August 2021.
\newblock arXiv:2007.04131 [stat].

\bibitem{callahan_interpretable_nodate}
Tiffany~J Callahan, Jie Shi, Kevin~J Cheng, Michael~A Sauer, Taras~V Pogorelov, and Sara Capponi.
\newblock Interpretable {Machine} {Learning} {Uncovers} {Structural} {Determinants} of {Wnt}-{Wls} {Binding} {Specificity} from {Extended} {Atomistic} {Simulations}.

\bibitem{lyons_predicting_2014}
James Lyons, Abdollah Dehzangi, Rhys Heffernan, Alok Sharma, Kuldip Paliwal, Abdul Sattar, Yaoqi Zhou, and Yuedong Yang.
\newblock Predicting backbone {Cα} angles and dihedrals from protein sequences by stacked sparse auto‐encoder deep neural network.
\newblock {\em Journal of Computational Chemistry}, 35(28):2040--2046, October 2014.

\bibitem{feig_is_2008}
Michael Feig.
\newblock Is {Alanine} {Dipeptide} a {Good} {Model} for {Representing} the {Torsional} {Preferences} of {Protein} {Backbones}?
\newblock {\em Journal of Chemical Theory and Computation}, 4(9):1555--1564, September 2008.

\bibitem{kikutsuji_explaining_2022}
Takuma Kikutsuji, Yusuke Mori, Kei-ichi Okazaki, Toshifumi Mori, Kang Kim, and Nobuyuki Matubayasi.
\newblock Explaining reaction coordinates of alanine dipeptide isomerization obtained from deep neural networks using {Explainable} {Artificial} {Intelligence} ({XAI}).
\newblock {\em The Journal of Chemical Physics}, 156(15):154108, April 2022.

\bibitem{kalousis_stability_2007}
Alexandros Kalousis, Julien Prados, and Melanie Hilario.
\newblock Stability of feature selection algorithms: a study on high-dimensional spaces.
\newblock {\em Knowledge and Information Systems}, 12(1):95--116, May 2007.

\bibitem{kursa_boruta_2010}
Miron~B. Kursa, Aleksander Jankowski, and Witold~R. Rudnicki.
\newblock Boruta – {A} {System} for {Feature} {Selection}.
\newblock {\em Fundamenta Informaticae}, 101(4):271--285, July 2010.

\bibitem{louppe_understanding_nodate}
Gilles Louppe, Louis Wehenkel, Antonio Sutera, and Pierre Geurts.
\newblock Understanding variable importances in forests of randomized trees.

\bibitem{jethani_fastshap_2022}
Neil Jethani, Mukund Sudarshan, Ian Covert, Su-In Lee, and Rajesh Ranganath.
\newblock {FastSHAP}: {Real}-{Time} {Shapley} {Value} {Estimation}, March 2022.
\newblock arXiv:2107.07436 [stat].

\bibitem{de_winter_comparing_2016}
Joost C.~F. De~Winter, Samuel~D. Gosling, and Jeff Potter.
\newblock Comparing the {Pearson} and {Spearman} correlation coefficients across distributions and sample sizes: {A} tutorial using simulations and empirical data.
\newblock {\em Psychological Methods}, 21(3):273--290, September 2016.

\bibitem{kuncheva_stability_nodate}
Ludmila~I Kuncheva.
\newblock A {STABILITY} {INDEX} {FOR} {FEATURE} {SELECTION}.

\bibitem{haury_influence_2011}
Anne-Claire Haury, Pierre Gestraud, and Jean-Philippe Vert.
\newblock The {Influence} of {Feature} {Selection} {Methods} on {Accuracy}, {Stability} and {Interpretability} of {Molecular} {Signatures}.
\newblock {\em PLoS ONE}, 6(12):e28210, December 2011.

\bibitem{wright_ranger_2017}
Marvin~N. Wright and Andreas Ziegler.
\newblock ranger: {A} {Fast} {Implementation} of {Random} {Forests} for {High} {Dimensional} {Data} in {C}++ and {R}.
\newblock {\em Journal of Statistical Software}, 77(1), 2017.
\newblock arXiv:1508.04409 [stat].

\bibitem{chen_xgboost_2016}
Tianqi Chen and Carlos Guestrin.
\newblock {XGBoost}: {A} {Scalable} {Tree} {Boosting} {System}.
\newblock In {\em Proceedings of the 22nd {ACM} {SIGKDD} {International} {Conference} on {Knowledge} {Discovery} and {Data} {Mining}}, pages 785--794, August 2016.
\newblock arXiv:1603.02754 [cs].

\bibitem{708428}
M.A. Hearst, S.T. Dumais, E.~Osuna, J.~Platt, and B.~Scholkopf.
\newblock Support vector machines.
\newblock {\em IEEE Intelligent Systems and their Applications}, 13(4):18--28, 1998.

\bibitem{beauchamp_are_2012}
Kyle~A. Beauchamp, Yu-Shan Lin, Rhiju Das, and Vijay~S. Pande.
\newblock Are {Protein} {Force} {Fields} {Getting} {Better}? {A} {Systematic} {Benchmark} on 524 {Diverse} {NMR} {Measurements}.
\newblock {\em Journal of Chemical Theory and Computation}, 8(4):1409--1414, April 2012.

\bibitem{piana_assessing_2014}
Stefano Piana, John~L Klepeis, and David~E Shaw.
\newblock Assessing the accuracy of physical models used in protein-folding simulations: quantitative evidence from long molecular dynamics simulations.
\newblock {\em Current Opinion in Structural Biology}, 24:98--105, February 2014.

\bibitem{ksenofontova_conformational_2013}
Olga Ksenofontova and Vasily Stefanov.
\newblock Conformational flexibility of the pharmacologically important insulin analogues.
\newblock {\em Advances in Biological Chemistry}, 03(05):512--517, 2013.

\bibitem{gewers_principal_2022}
Felipe~L. Gewers, Gustavo~R. Ferreira, Henrique F.~de Arruda, Filipi~N. Silva, Cesar~H. Comin, Diego~R. Amancio, and Luciano da~F. Costa.
\newblock Principal {Component} {Analysis}: {A} {Natural} {Approach} to {Data} {Exploration}.
\newblock {\em ACM Computing Surveys}, 54(4):1--34, May 2022.
\newblock arXiv:1804.02502 [cs].

\end{thebibliography}

\appendix 

\section{Software Availability}
This work also contributes a reproducible R package, \href{https://cloud.r-project.org/web/packages/TangledFeatures/index.html}{\texttt{TangledFeatures}}
, available on GitHub and prepared for CRAN submission. The package implements the core functions for redundancy detection, pruning, and stability analysis, along with visualization utilities for correlation structures. Documentation includes example workflows across multiple domains, and extensions are underway to handle high-dimensional genomics, complex financial markets, and large-scale government datasets. By making the tool openly available, we aim to encourage adoption by both researchers and practitioners seeking interpretable and redundancy-aware feature selection methods. 

\begin{figure}[ht]
    \centering
    \subcaption*{\textbf{Feature selection stability for $\phi$ torsional angle}}
    
    \begin{subfigure}[t]{0.45\linewidth}
        \includegraphics[width=\linewidth]{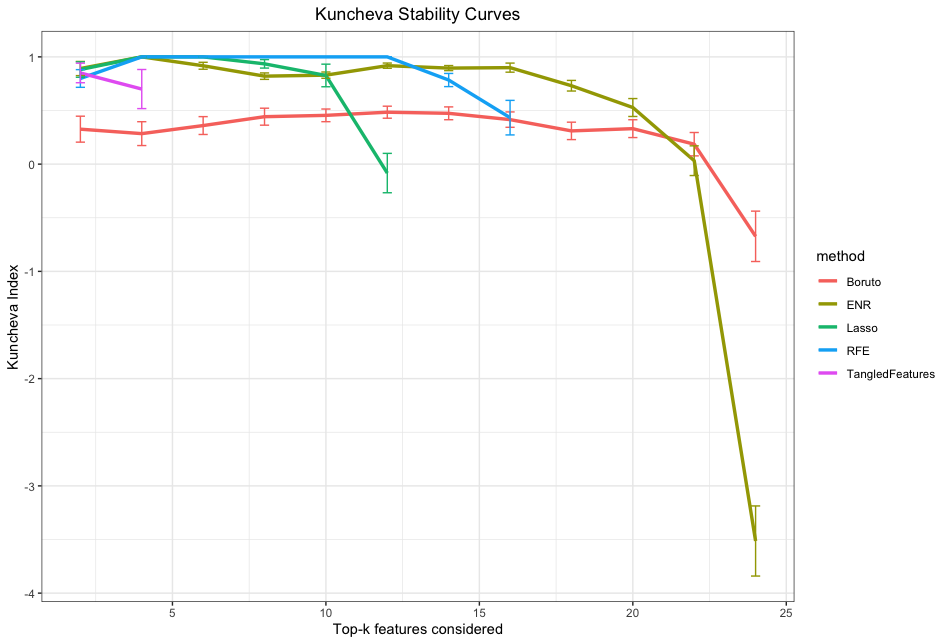}
        \caption{OLS: Kuncheva}
    \end{subfigure}
    \hfill
    \begin{subfigure}[t]{0.45\linewidth}
        \includegraphics[width=\linewidth]{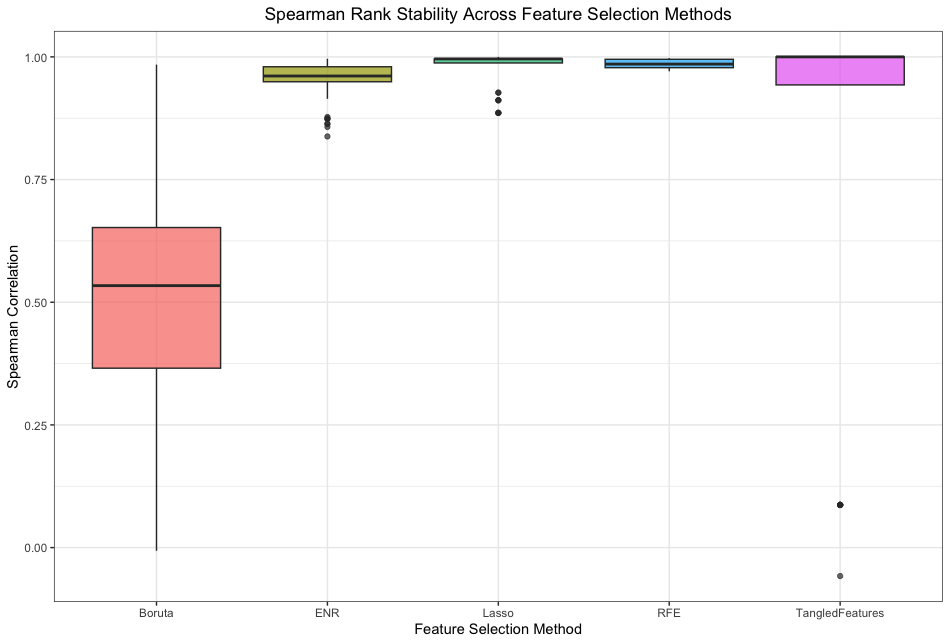}
        \caption{OLS: Spearman}
    \end{subfigure}
    
    \begin{subfigure}[t]{0.45\linewidth}
        \includegraphics[width=\linewidth]{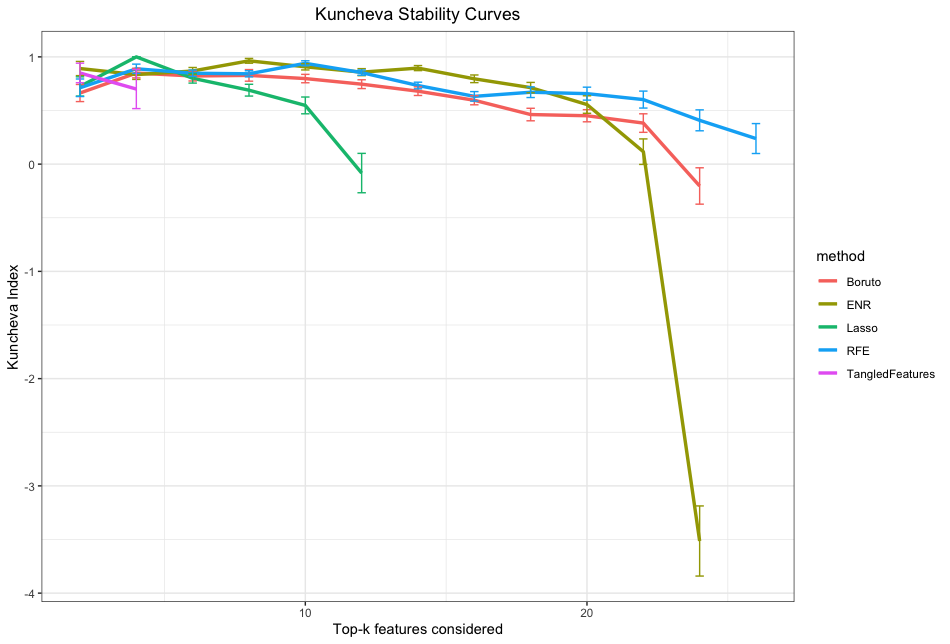}
        \caption{SVM: Kuncheva}
    \end{subfigure}
    \hfill
    \begin{subfigure}[t]{0.45\linewidth}
        \includegraphics[width=\linewidth]{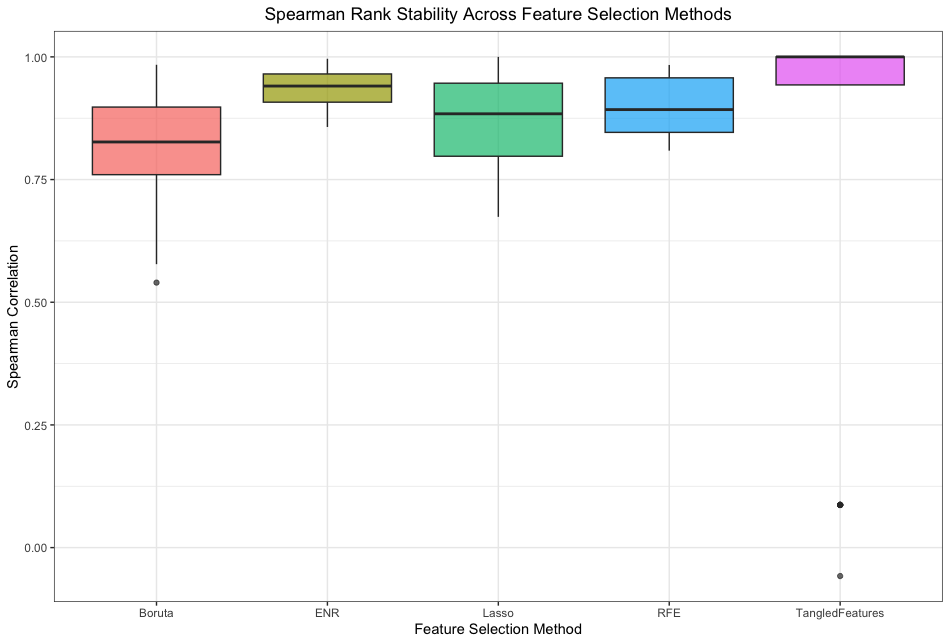}
        \caption{SVM: Spearman}
    \end{subfigure}
    
    \begin{subfigure}[t]{0.45\linewidth}
        \includegraphics[width=\linewidth]{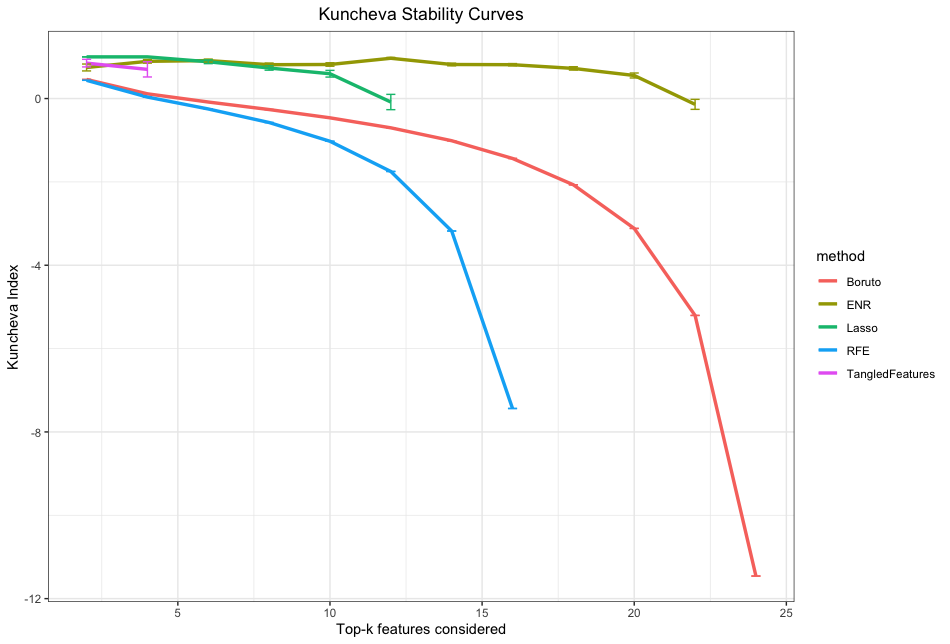}
        \caption{XGBoost: Kuncheva}
    \end{subfigure}
    \hfill
    \begin{subfigure}[t]{0.45\linewidth}
        \includegraphics[width=\linewidth]{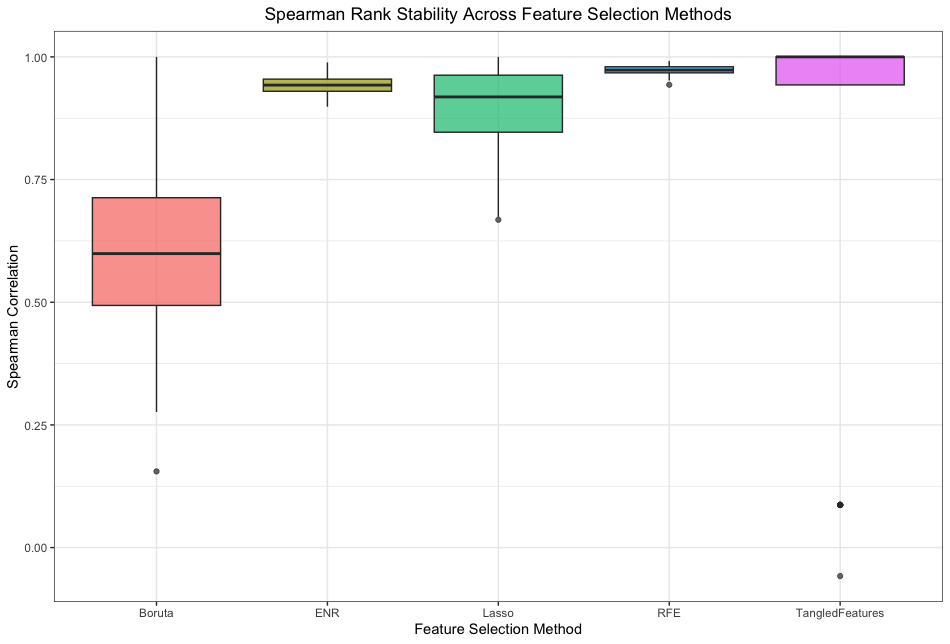}
        \caption{XGBoost: Spearman}
    \end{subfigure}
    
    \begin{subfigure}[t]{0.45\linewidth}
        \includegraphics[width=\linewidth]{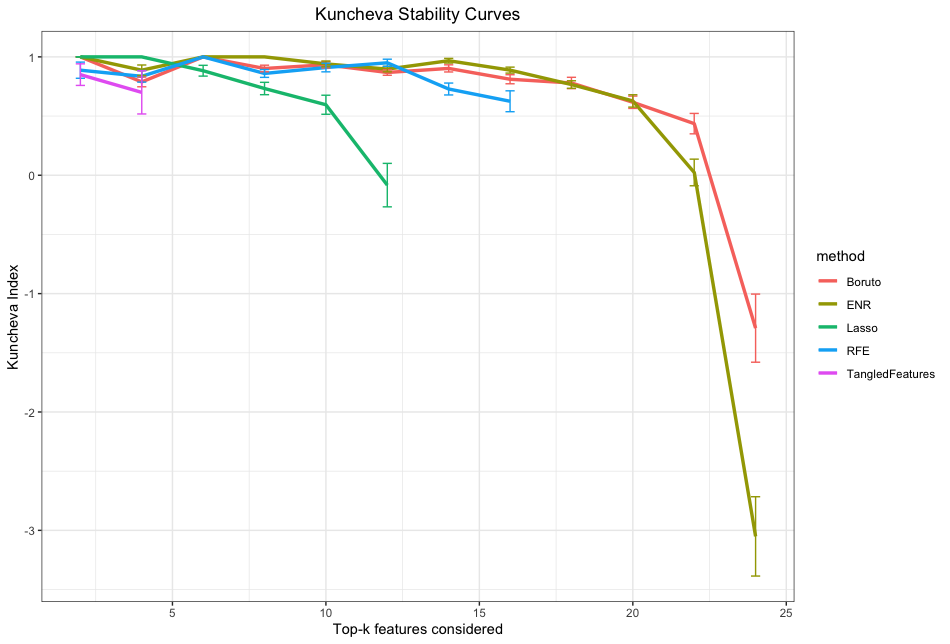}
        \caption{RF: Kuncheva}
    \end{subfigure}
    \hfill
    \begin{subfigure}[t]{0.45\linewidth}
        \includegraphics[width=\linewidth]{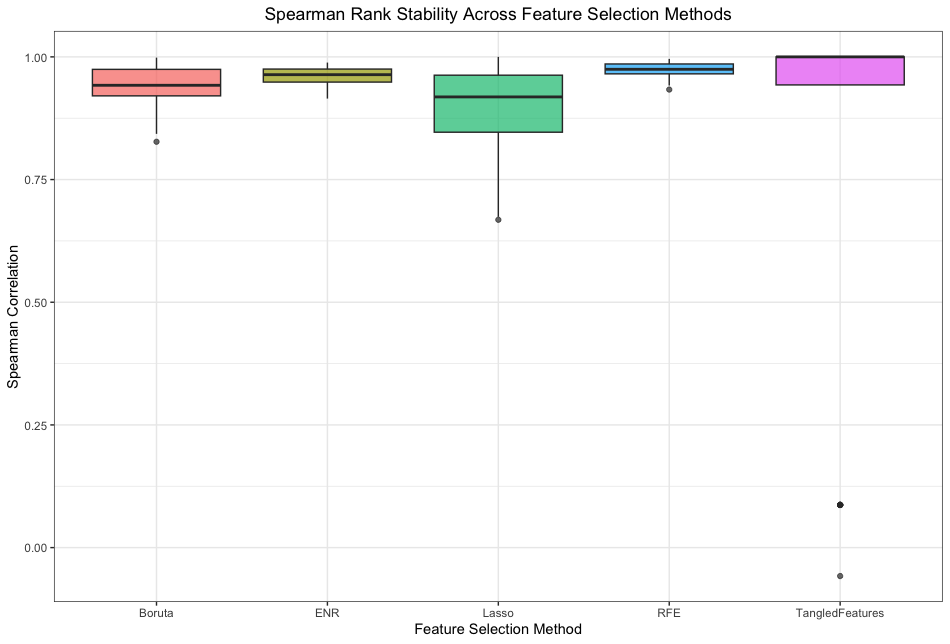}
        \caption{RF: Spearman}
    \end{subfigure}
    
    \caption{Stability curves (Kuncheva index, left column; Spearman rank correlation, right column) for $\phi$ torsional angle across predictive models (rows: OLS, SVM, XGBoost, RF).}
    \label{fig:phi_stability_grid}
\end{figure}

\begin{figure}[ht]
    \centering
    \subcaption*{\textbf{Feature selection stability for $\psi$ torsional angle}}
    
    \begin{subfigure}[t]{0.45\linewidth}
        \includegraphics[width=\linewidth]{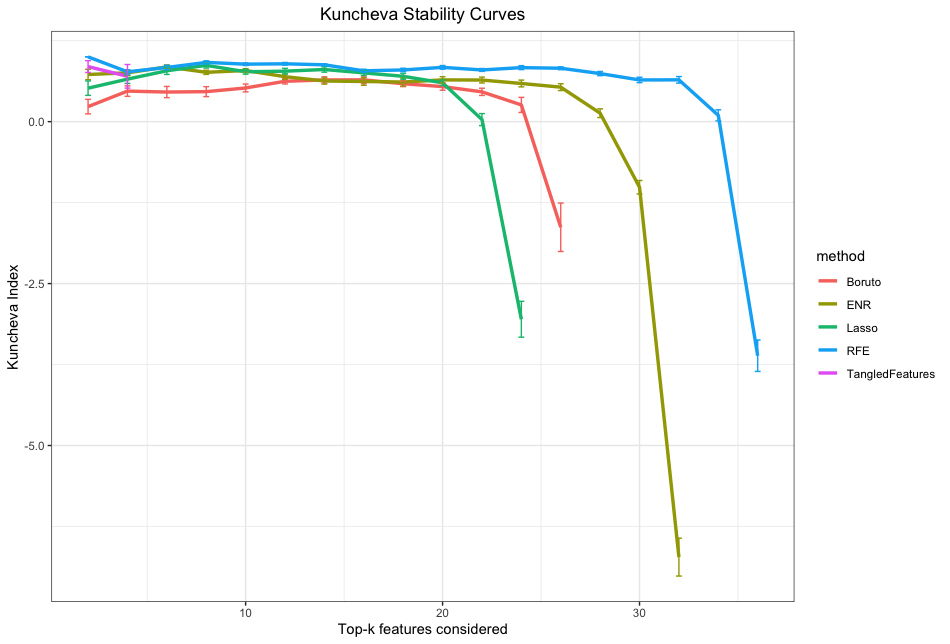}
        \caption{OLS: Kuncheva}
    \end{subfigure}
    \hfill
    \begin{subfigure}[t]{0.45\linewidth}
        \includegraphics[width=\linewidth]{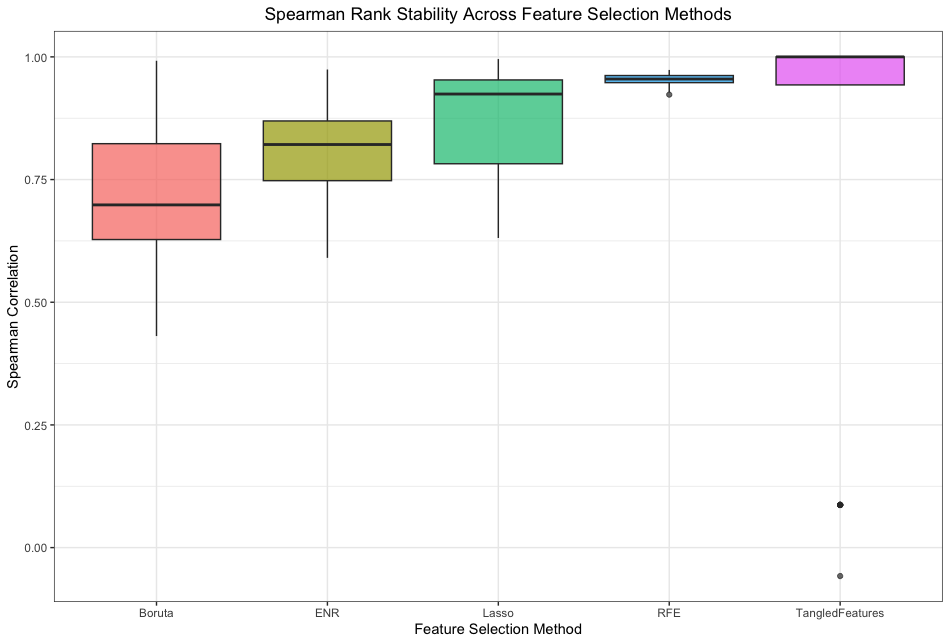}
        \caption{OLS: Spearman}
    \end{subfigure}
    
    \begin{subfigure}[t]{0.45\linewidth}
        \includegraphics[width=\linewidth]{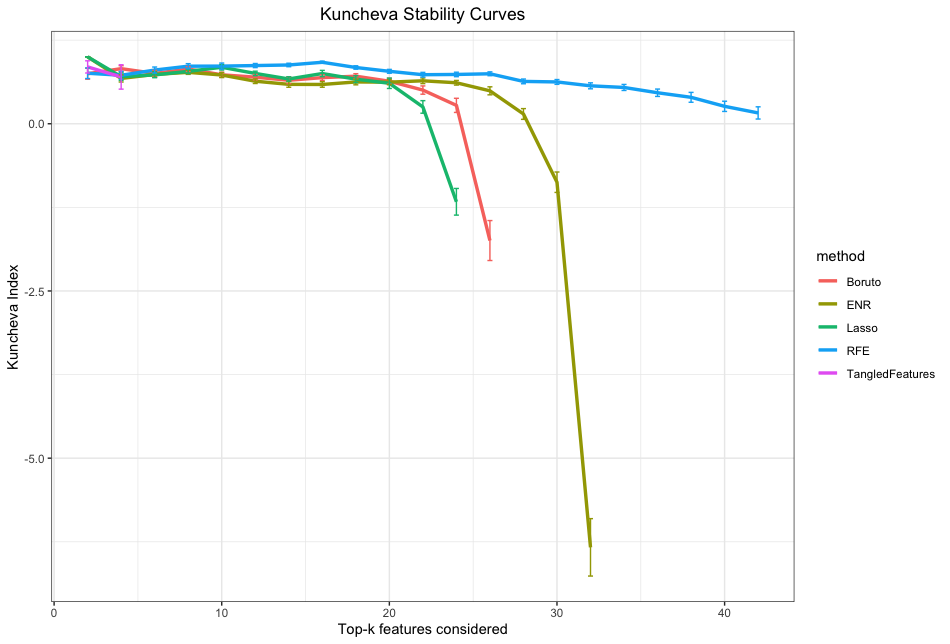}
        \caption{SVM: Kuncheva}
    \end{subfigure}
    \hfill
    \begin{subfigure}[t]{0.45\linewidth}
        \includegraphics[width=\linewidth]{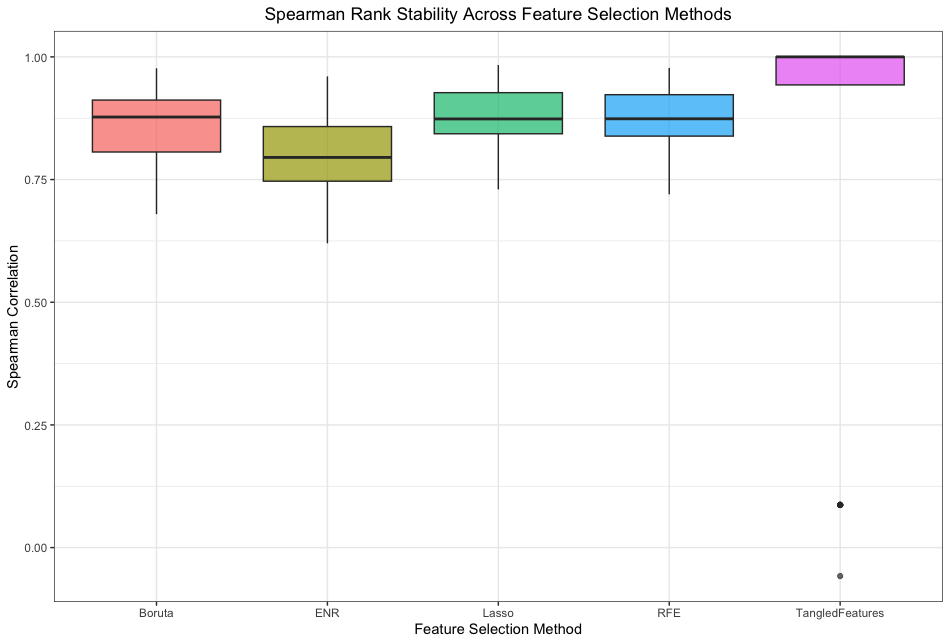}
        \caption{SVM: Spearman}
    \end{subfigure}
    
    \begin{subfigure}[t]{0.45\linewidth}
        \includegraphics[width=\linewidth]{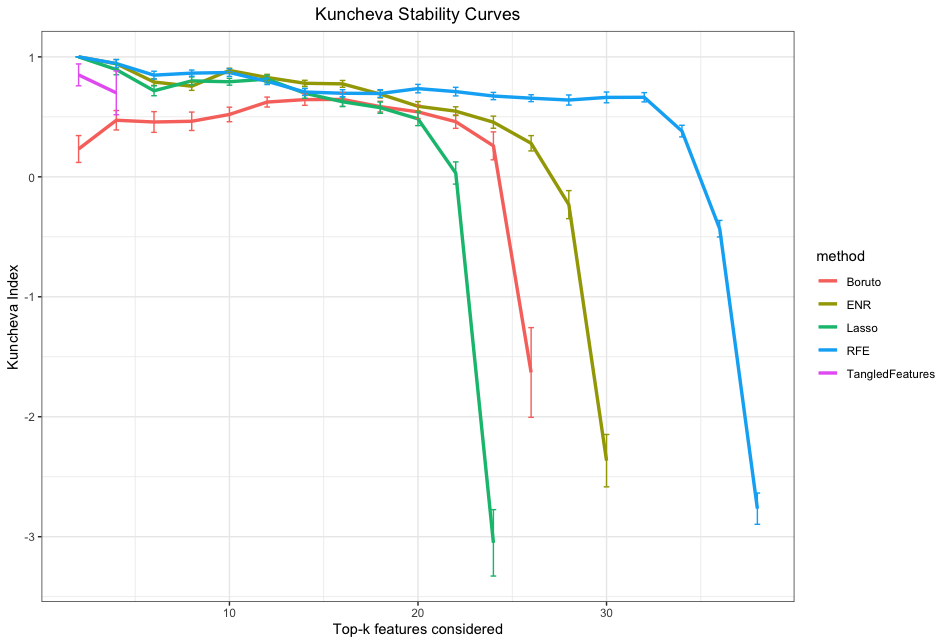}
        \caption{XGBoost: Kuncheva}
    \end{subfigure}
    \hfill
    \begin{subfigure}[t]{0.45\linewidth}
        \includegraphics[width=\linewidth]{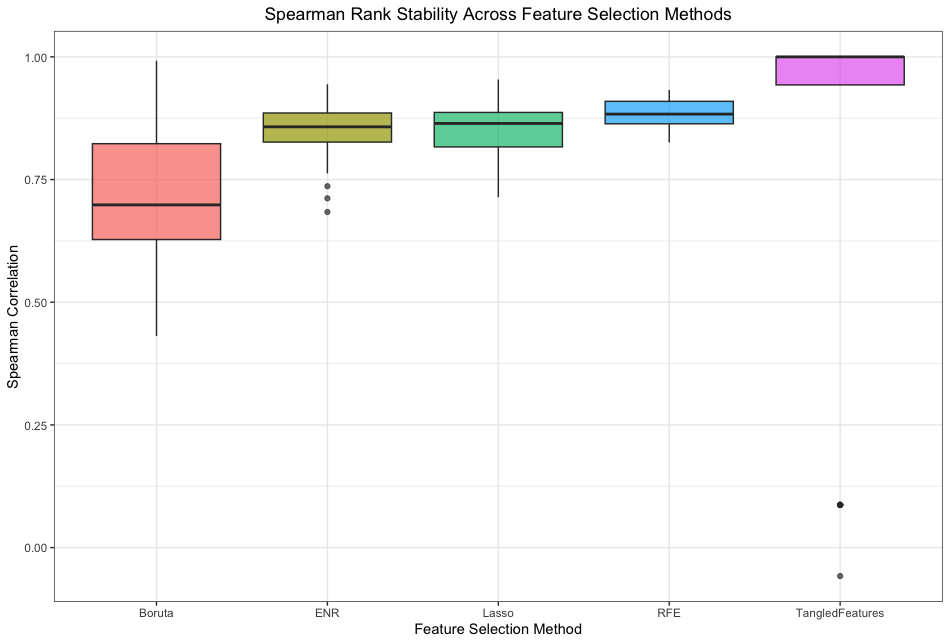}
        \caption{XGBoost: Spearman}
    \end{subfigure}
    
    \begin{subfigure}[t]{0.45\linewidth}
        \includegraphics[width=\linewidth]{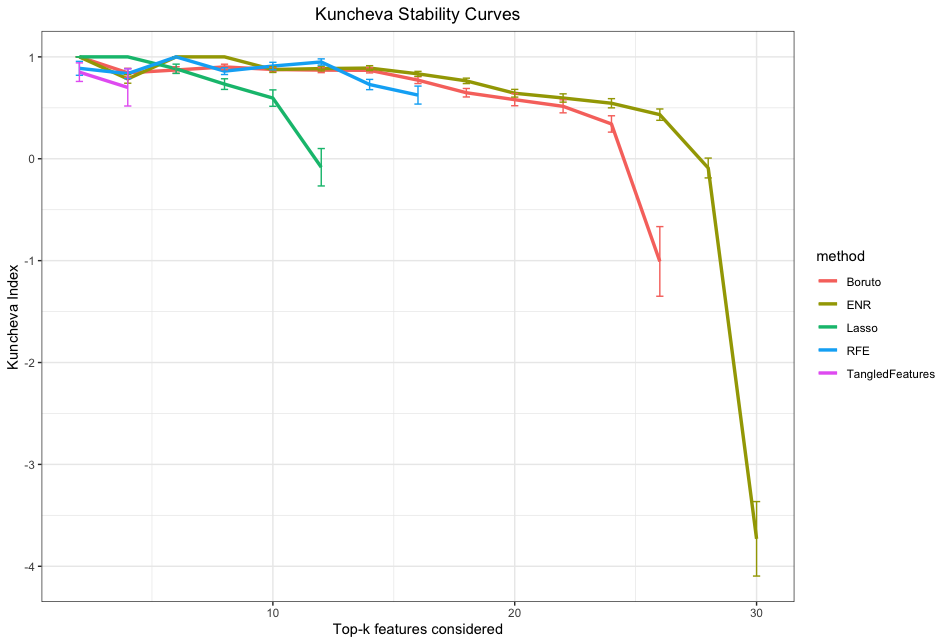}
        \caption{RF: Kuncheva}
    \end{subfigure}
    \hfill
    \begin{subfigure}[t]{0.45\linewidth}
        \includegraphics[width=\linewidth]{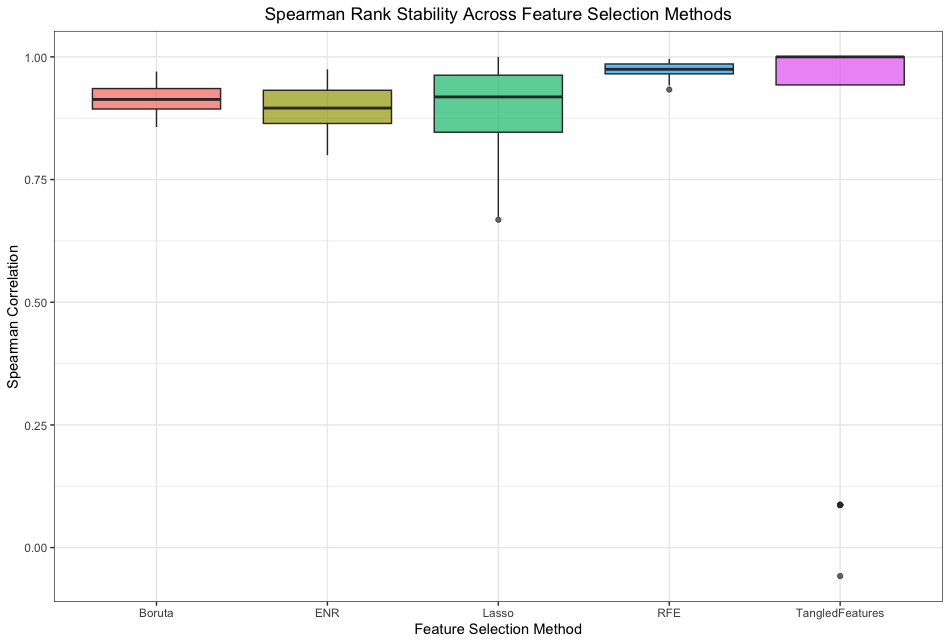}
        \caption{RF: Spearman}
    \end{subfigure}
    
    \caption{Stability curves (Kuncheva index, left column; Spearman rank correlation, right column) for $\psi$ torsional angle across predictive models (rows: OLS, SVM, XGBoost, RF).}
    \label{fig:psi_stability_grid}
\end{figure}

\section{Selected Distances} \label{appendix:selected-distances}

Table~\ref{tab:tangled_atoms_phi} and Table~\ref{tab:tangled_atoms_psi} list the intra-atomic distances retained by TangledFeatures after correlation pruning and refinement. These represent the compact subsets ultimately selected for $\phi$ and $\psi$, respectively by TangledFeatures.

\begin{table}[H]
\centering
\begin{tabular}{ll}
\toprule
\textbf{Atom Distance} & \textbf{Category} \\
\midrule
ACE1-CH3 $\leftrightarrow$ ALA2-CB & Cap-related/Flexible \\
ACE1-C $\leftrightarrow$ ALA2-C   & Flexible \\
ACE1-CH3 $\leftrightarrow$ ALA2-CA & Flexible \\
ACE1-C $\leftrightarrow$ ACE1-O   & Rigid \\
ACE1-CH3 $\leftrightarrow$ NME3-N & Cap-related \\
ACE1-C $\leftrightarrow$ NME3-N   & Cap-related \\
ACE1-CH3 $\leftrightarrow$ ALA2-C & Rigid \\
\bottomrule
\end{tabular}
\vspace{0.5em}
\caption{Unique atom--atom distances selected by TangledFeatures for $\phi$ and their categorical assignments.}
\label{tab:tangled_atoms_phi}
\end{table}

\begin{table}[H]
\centering
\begin{tabular}{ll}
\toprule
\textbf{Atom Distance} & \textbf{Category} \\
\midrule
ACE1-CH3 $\leftrightarrow$ ALA2-C   & Rigid \\
ACE1-C $\leftrightarrow$ NME3-N     & Cap-related \\
ACE1-C $\leftrightarrow$ ACE1-O     & Rigid \\
ACE1-C $\leftrightarrow$ ALA2-O     & Rigid \\
ACE1-CH3 $\leftrightarrow$ NME3-N   & Cap-related \\
ACE1-C $\leftrightarrow$ ALA2-C     & Flexible \\
\bottomrule
\end{tabular}
\vspace{0.5em}
\caption{Unique atom--atom distances selected by TangledFeatures for $\psi$ and their categorical assignments.}
\label{tab:tangled_atoms_psi}
\end{table}

\end{document}